\documentclass[10pt,twocolumn,letterpaper]{article}

\usepackage[utf8]{inputenc}
\usepackage[T1]{fontenc}
\usepackage{cvpr}
\usepackage{times}
\usepackage{epsfig}
\usepackage{graphicx}
\usepackage{amsmath}
\usepackage{amssymb}
\usepackage{textcomp}
\usepackage{gensymb}

\usepackage{color}

\usepackage{booktabs,tabu}

\DeclareMathOperator*{\argmin}{arg\,min}

\usepackage[pagebackref=true,breaklinks=true,letterpaper=true,colorlinks,bookmarks=false]{hyperref}

\cvprfinalcopy 

\def\cvprPaperID{3565} 

\ifcvprfinal\fi
\begin{document}

\title{Deep Outdoor Illumination Estimation}

\author{Yannick Hold-Geoffroy\textsuperscript{1*},\: Kalyan Sunkavalli\textsuperscript{$\dagger$},\: Sunil Hadap\textsuperscript{$\dagger$},\: Emiliano Gambaretto\textsuperscript{$\dagger$},\: Jean-Fran\c{c}ois Lalonde\textsuperscript{*}\\
Universit\'e Laval\textsuperscript{*}, Adobe Research\textsuperscript{$\dagger$}\\
{\tt\small yannick.hold-geoffroy.1@ulaval.ca, \{sunkaval,hadap,emiliano\}@adobe.com, jflalonde@gel.ulaval.ca}\\
\small\url{http://www.jflalonde.ca/projects/deepOutdoorLight}}

\maketitle

\begin{abstract}
We present a CNN-based technique to estimate high-dynamic range outdoor illumination from a single low dynamic range image. To train the CNN, we leverage a large dataset of outdoor panoramas. We fit a low-dimensional physically-based outdoor illumination model to the skies in these panoramas giving us a compact set of parameters (including sun position, atmospheric conditions, and camera parameters). We extract limited field-of-view images from the panoramas, and train a CNN with this large set of input image--output lighting parameter pairs. Given a test image, this network can be used to infer illumination parameters that can, in turn, be used to reconstruct an outdoor illumination environment map. We demonstrate that our approach allows the recovery of plausible illumination conditions and enables photorealistic virtual object insertion from a single image. An extensive evaluation on both the panorama dataset and captured HDR environment maps shows that our technique significantly outperforms previous solutions to this problem.

\end{abstract}

\makeatletter
\def\blfootnote{\gdef\@thefnmark{}\@footnotetext}
\makeatother

\blfootnote{\scriptsize \textsuperscript{1} Research partly done when Y. Hold-Geoffroy was an intern at Adobe Research.}

\section{Introduction}

\begin{figure}
\centering
\includegraphics[width=\linewidth]{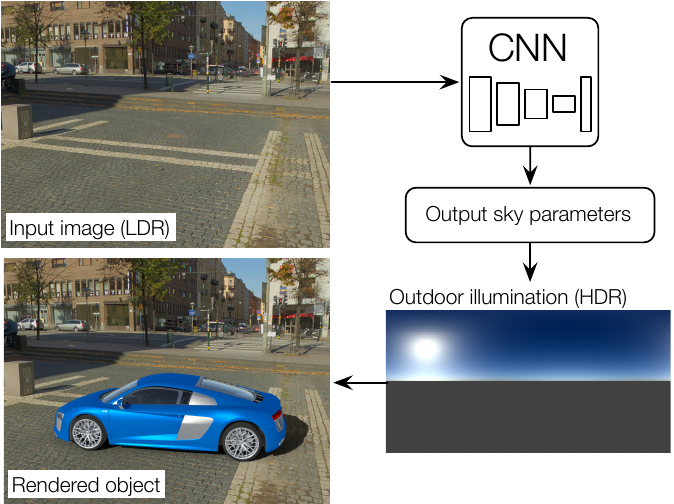}
\caption{We present an approach for predicting full HDR lighting conditions from a single LDR outdoor image. Our prediction can readily be used to insert a virtual object into the image. Our key idea is to train a CNN using input-output pairs of LDR images and HDR illumination parameters that are automatically extracted from a large database of $360^\circ$ panoramas.}
\label{fig:teaser}
\vspace{-1em}
\end{figure}

Illumination plays a critical role in deciding the appearance of a scene, and recovering scene illumination is important for a number of tasks ranging from scene understanding to reconstruction and editing. However, the process of image formation conflates illumination with scene geometry and material properties in complex ways and inverting this process is an extremely ill-posed problem. This is especially true in outdoor scenes, where we have little to no control over the capture process.

Previous approaches to this problem have relied on extracting cues such as shadows and shading~\cite{lalonde-ijcv-12} and combining them with (reasonably good) estimates of scene geometry to recover illumination. However, both these tasks are challenging and existing attempts often result in poor performance on real-world images. Alternatively, techniques for intrinsic images can estimate low-frequency illumination but rely on hand-tuned priors on geometry and material properties~\cite{barron-pami-15,lombardi2016reflectance} that may not generalize to large-scale scenes. In this work, we seek a single image outdoor illumination inference technique that generalizes to a wide range of scenes and does not make strong assumptions about scene properties.

To this end, our goal is to train a CNN to directly regress a single input low dynamic range image to its corresponding high dynamic range (HDR) outdoor lighting conditions. Given the success of deep networks at related tasks like intrinsic images~\cite{zhou2015intrinsic} and reflectance map estimation~\cite{rematas-cvpr-16}, our hope is that an appropriately designed CNN can learn this relationship. However, training such a CNN requires a very large dataset of outdoor images with their corresponding HDR lighting conditions. Unfortunately, such a dataset currently does not exist, and, because capturing light probes requires significant time and effort, acquiring it is prohibitive.

Our insight is to exploit a large dataset of outdoor panoramas~\cite{xiao-cvpr-12}, and extract photos with limited field of view from them. We can thus use pairs of photos and panoramas to train the neural network. However, this approach is bound to fail since: 1) the panoramas have low dynamic range and therefore do not provide an accurate estimate of outdoor lighting; and 2) even if notable attempts have been made~\cite{zhang-cvpr-13}, recovering full spherical panoramas from a single photo is both improbable and unnecessary for a number of tasks (e.g., many of the high-frequency details in the panoramas are not required when rendering Lambertian objects into the scene).

Instead, we use a physically-based sky model---the Ho\v{s}ek-Wilkie model~\cite{hosek-siggraph-12,hosek-cga-13}---and fit its parameters to the visible sky regions in the input panorama. This has two advantages: first, it allows us to recover physically accurate, high dynamic range information from the panoramas (even in saturated regions). Second, it compresses the panorama to a compact set of physically meaningful and representative parameters that can be efficiently learned by a CNN. At test time, we recover these parameters---including sun position, atmospheric turbidity, and geometric and radiometric camera calibration---from an input image and use them to construct an HDR sky environment map. 

To our knowledge, we are the first to address the complete scope of estimating a full HDR lighting representation---which can readily be used for image-based lighting~\cite{Debevec1998}---from a single outdoor image (fig.~\ref{fig:teaser}). Previous techniques have typically addressed only aspects of this problem, e.g., Lalonde et al.~\cite{lalonde-ijcv-12} recover the position of the sun but need to observe sky pixels in order to recover the atmospheric conditions. Similarly,~\cite{Ma2017} uses a neural network to estimate the sun azimuth to perform localization in roadside environments. Karsch et al.~\cite{karsch2014automatic} estimate full environment map lighting, but their panorama transfer technique may yield illumination conditions arbitrarily far away from the real ones. In contrast, our technique can recover an accurate, full HDR sky environment map from an arbitrary input image. We show through extensive evaluation that our estimates of the lighting conditions are significantly better than previous techniques and that they can be used ``as is'' to photorealistically relight and render 3D models into images.

\section{Related work}

\paragraph{Outdoor illumination models} Perez et al.~\cite{perez1993allweather} proposed an all-weather sky luminance distribution model. This model was a generalization of the CIE standard sky model and is parameterized by five coefficients that can be varied to generate a wide range of skies. Preetham~\cite{preetham-siggraph-99} proposed a simplified version of the Perez model that explains the five coefficients using a single unified atmospheric turbidity parameter. Lalonde and Matthews~\cite{lalonde-3dv-14} combined the Preetham sky model with a novel empirical sun model. Ho\v{s}ek and Wilkie proposed a sky luminance model~\cite{hosek-siggraph-12} and solar radiance function~\cite{hosek-cga-13}.

\vspace{-0.5em}
\paragraph{Outdoor lighting estimation} Lalonde et al.~\cite{lalonde-ijcv-12} combine multiple cues, including shadows, shading of vertical surfaces, and sky appearance to predict the direction and visibility of the sun. This is combined with an estimation of sky illumination (represented by the Perez model~\cite{perez1993allweather}) from sky pixels~\cite{lalonde-ijcv-10}. Similar to this work, we use a physically-based model for outdoor illumination. However, instead of designing hand-crafted features to estimate illumination, we train a CNN to directly learn the highly complex mapping between image pixels and illumination parameters. 

Other techniques for single image illumination estimation rely on known geometry and/or strong priors on scene reflectance, geometry and illumination~\cite{barron-pami-15,barron2013rgbd,lombardi2016reflectance}. These priors typically do not generalize to large-scale outdoor scenes. Karsch et al.~\cite{karsch2014automatic} retrieve panoramas (from the SUN360 panorama dataset~\cite{xiao-cvpr-12}) with features similar to the input image, and refine the retrieved panoramas to compute the illumination. However, the matching metric is based on image content which may not be directly linked with illumination. 

Another class of techniques simplify the problem by estimating illumination from image collections. Multi-view image collections have been used to reconstruct geometry, which is used to recover outdoor illumination~\cite{haber2009relighting,lalonde-3dv-14,shan2015visual,duchene2015multiview}, sun direction~\cite{wehrwein2015shadows}, or place and time of capture~\cite{hauagge2014outdoor}. Appearance changes have also been used to recover colorimetric variations of outdoor sun-sky illumination~\cite{sunkavalli2008color}. 

\vspace{-0.5em}
\paragraph{Inverse graphics/vision problems in deep learning} Following the remarkable success of deep learning-based methods on high-level recognition problems, these approaches are now being increasingly used to solve inverse graphics problems~\cite{kulkarni15dcign}. In the context of understanding scene appearance, previous work has leveraged deep learning to estimate depth and surface normals~\cite{eigen2015depth,bansal2016marr}, recognize materials~\cite{bell2015minc}, decompose intrinsic images~\cite{zhou2015intrinsic}, recover reflectance maps~\cite{rematas-cvpr-16}, and estimate, in a setup similar to physics-based techniques~\cite{lombardi2016reflectance}, lighting from objects of specular materials~\cite{georgoulis2016delight}. We believe ours is the first attempt at using deep learning for full HDR outdoor lighting estimation from a single image.

\section{Overview}

We aim to train a CNN to predict illumination conditions from a single outdoor image. We use full spherical, 360\degree ~panoramas, as they capture scene appearance while also providing a direct view of the sun and sky, which are the most important sources of light outdoors. Unfortunately, there exists no database containing true high dynamic range outdoor panoramas, and we must resort to using the saturated, low dynamic range panoramas in the SUN360 dataset~\cite{xiao-cvpr-12}. To overcome this limitation, and to provide a small set of meaningful parameters to learn to the CNN, we first fit a physically-based sky model to the panoramas (sec.~\ref{sec:dataset}). Then, we design and train a CNN that given an input image sampled from the panorama, outputs the fit illumination parameters (sec.~\ref{sec:cnn}), and thoroughly evaluate its performance in sec.~\ref{sec:evaluation}.

Throughout this paper, and following~\cite{xiao-cvpr-12}, will use the term \emph{photo} to refer to a standard limited-field-of-view image as taken with a normal camera, and the term \emph{panorama} to denote a 360-degree full-view panoramic image.
\section{Dataset preparation}
\label{sec:dataset}

\begin{figure}
\centering
\footnotesize
\setlength{\tabcolsep}{1pt}
\begin{tabular}{cccc} 
\includegraphics[width=.24\linewidth]{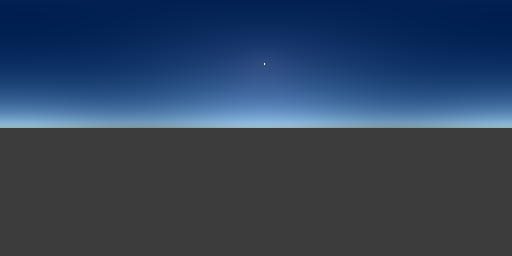} &
\includegraphics[width=.24\linewidth]{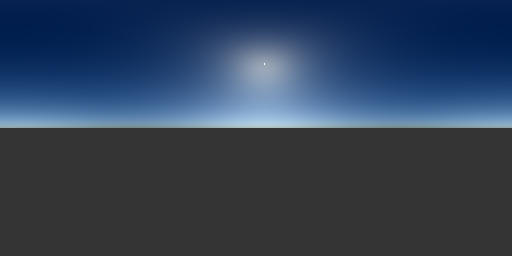} &
\includegraphics[width=.24\linewidth]{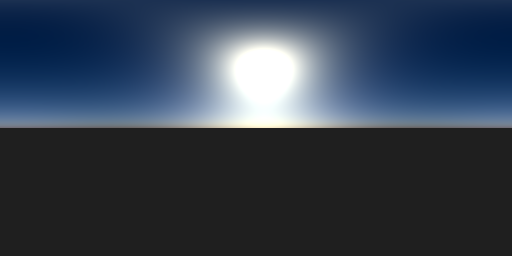} &
\includegraphics[width=.24\linewidth]{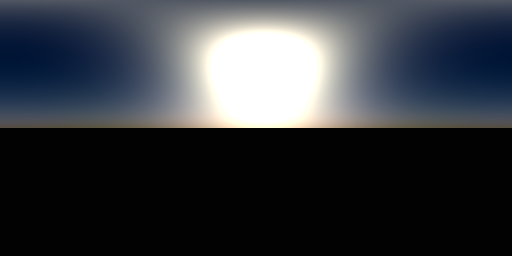} \\
\includegraphics[width=.24\linewidth]{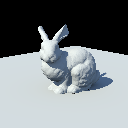} &
\includegraphics[width=.24\linewidth]{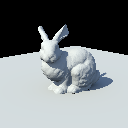} &
\includegraphics[width=.24\linewidth]{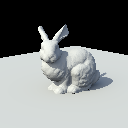} &
\includegraphics[width=.24\linewidth]{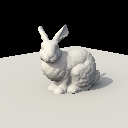} \\
$t = 1$ & $t = 3$ & $t = 7$ & $t = 10$ 
\end{tabular}
\caption{Impact of sky turbidity $t$ on rendered objects. The top row shows environment maps (in latitude-longitude format), and the bottom row shows corresponding renders of a bunny model on a ground plane for varying values for the turbidity $t$, ranging from low (left) to high (right). Images have been tonemapped with $\gamma = 2.2$ for display.}
\label{fig:turbidity-comparison}
\end{figure}

In this section, we detail the steps taken to augment the SUN360 dataset~\cite{xiao-cvpr-12} with HDR data via the use of the Ho\v{s}ek-Wilkie sky model, and simultaneously extract lighting parameters that can be learned by the network. We first briefly describe the sky model parameterization, followed by the optimization strategy used to recover its parameters from a LDR panorama. 

\subsection{Sky lighting model}

We employ the model proposed by Ho\v{s}ek and Wilkie~\cite{hosek-siggraph-12}, which has been shown~\cite{kider-tog-14} to more accurately represent skylight than the popular Preetham model~\cite{preetham-siggraph-99}. The model has also been extended to include a solar radiance function~\cite{hosek-cga-13}, which we also exploit.

In its simplest form, the Ho\v{s}ek-Wilkie (HW) model expresses the spectral radiance $L_\lambda$ of a lighting direction along the sky hemisphere $\mathbf{l} \in \Omega_\text{sky}$ as a function of several parameters:
\begin{equation}
L_\lambda(\mathbf{l}) = f_\text{HW}(\mathbf{l}, \lambda, t, \sigma_g, \mathbf{l}_s) \,,
\end{equation}
where $\lambda$ is the wavelength, $t$ the atmospheric turbidity (a measure of the amount of aerosols in the air), $\sigma_g$ the ground albedo, and $\mathbf{l}_s$ the sun position. Here, we fix $\sigma_g = 0.3$ (approximate average albedo of the Earth~\cite{goode-grl-01}). 

From this spectral model, we obtain RGB values by rendering it at a discrete set wavelengths spanning the 360--700nm spectrum, convert to CIE XYZ via the CIE standard observer color matching functions, and finally convert again from XYZ to CIE RGB~\cite{hosek-siggraph-12}. Referring to this conversion process as $f_\text{RGB}(\cdot)$, we express the RGB color $C_\text{RGB}(\mathbf{l})$ of a sky direction $\mathbf{l}$ as the following expression:
\begin{equation}
C_\text{RGB}(\mathbf{l}) = \omega f_\text{RGB}(\mathbf{l}, t, \mathbf{l}_s)\,.
\label{eqn:rgb-model} 
\end{equation}

In this equation, $\omega$ is a scale factor applied to all three color channels, which aims at estimating the (arbitrary and varying) exposure for each panorama. To generate a sky environment map from this model, we simply discretize the sky hemisphere $\Omega_\text{sky}$ into several directions (in this paper, we use the latitude-longitude format~\cite{reinhard-book-10}), and render the RGB values with (\ref{eqn:rgb-model}). Pixels which fall within $0.25^\circ$ of the sun position $\mathbf{l}_s$ are rendered with the HW sun model~\cite{hosek-cga-13} instead (converted to RGB as explained above).

Thus, we are left with three important parameters: the sun position $\mathbf{l}_s$, which indicate where the main directional light source is located in the sky, the exposure $\omega$, and the turbidity $t$. The turbidity is of paramount importance as it controls the relative sun color (and intensity) with respect to that of the sky. As illustrated in fig.~\ref{fig:turbidity-comparison}, a low turbidity indicates a clear sky with a very bright sun, and a high turbidity represents a sky closer that is closer to overcast situations, where the sun is much dimmer. 

\subsection{Optimization procedure}
\label{sec:optimization}

We now describe how the sky model parameters are estimated from a panorama in the SUN360 dataset. This procedure is carefully crafted to be robust to the extremely varied set of conditions encountered in the dataset which severely violates the linear relationship between sky radiance and pixel values such as: unknown camera response function and white-balance, manual post-processing by photographers and stitching artifacts. 

Given a panorama $P$ in latitude-longitude format and a set of pixels indices $p \in \mathcal{S}$ corresponding to sky pixels in $P$, we wish to obtain the sun position $\mathbf{l}_s$, exposure $\omega$ and sky turbidity $t$ by minimizing the visible sky reconstruction error in a least-squares sense: 
\begin{equation}
\begin{split}
\mathbf{l}_s^*,\omega^*,t^* =& \argmin_{\mathbf{l}_s,\omega,t} \sum_{p \in \Omega_s} \left(P(p)^\gamma - \omega f_\text{RGB}(\mathbf{l}_p, t, \mathbf{l}_s) \right)^2 \\
& \text{s.t.} \quad t \in [1, 10] \,, \label{eq:optim_1omega}
\end{split}
\end{equation}
where $f_\text{RGB}(\cdot)$ is defined in (\ref{eqn:rgb-model}) and $\mathbf{l}_p$ is the light direction corresponding to pixel $p \in \Omega_s$ (according to the latitude-longitude mapping). Here, we model the inverse response function of the camera with a simple gamma curve ($\gamma = 2.2$). Optimizing for gamma was found to be unstable and keeping it fixed yielded much more robust results. 
 
We solve (\ref{eq:optim_1omega}) in a 2-step procedure. First, the sun position $\mathbf{l}_s$ is estimated by finding the largest connected component of the sky above a threshold (98th percentile), and by computing its centroid. The sun position is fixed at this value, as it was determined that optimizing for its position at the next stage too often made the algorithm converge to undesirable local minima. 

Second, the turbidity $t$ is initialized to $\{1, 2, 3, ..., 10\}$ and (\ref{eq:optim_1omega}) is optimized using the Trust Region Reflective algorithm (a variant of the Levenberg-Marquardt algorithm which supports bounds) for each of these starting points. The parameters resulting in the lowest error are kept as the final result. During the optimization loop, for the current value of $t$, $\omega^*$ is obtained through the closed-form solution
\begin{equation}
\label{eq:omega_cfs}
\omega^* = \frac{\sum_{p \in \mathcal{S}} P(p) f_\text{RGB}(\mathbf{l}_p, t, \mathbf{l}_{s})}{\sum_{p \in \mathcal{S}} f_\text{RGB}(\mathbf{l}_p, t, \mathbf{l}_s)^2} \,.
\end{equation}
Finally, the sky mask $\mathcal{S}$ is obtained with the sky segmentation method of~\cite{tsai-siggraph-16}, followed by a CRF refinement~\cite{krahenbuhl-nips-12}.

\begin{figure}
\centering
\begin{tabu} to 7cm {lX[c]X[c]}
\toprule
\textbf{Layer} & \textbf{Stride} & \textbf{Resolution} \\
\midrule
Input & & $320 \times 240$ \\
\midrule
conv7-64  & 2 & $160 \times 120$ \\
conv5-128 & 2 & $80 \times 60$ \\
conv3-256 & 2 & $40 \times 30$ \\
conv3-256 & 1 & $40 \times 30$ \\
conv3-256 & 2 & $20 \times 15$ \\
conv3-256 & 1 & $20 \times 15$ \\
conv3-256 & 2 & $10 \times 8$ \\
\midrule
\multicolumn{3}{c}{FC-2048} \\
\midrule
\end{tabu} \\
\begin{tabu} to 7cm {X[c]X[c]}
FC-160 & FC-5 \\
LogSoftMax & Linear \\*[-.5em]
\noindent\rule{3.4cm}{.8pt} &
\noindent\rule{3.4cm}{.8pt} \\
Output: sun position distribution $\mathbf{s}$ &
Output: sky and camera parameters $\mathbf{q}$ \\
\end{tabu}
\vspace{.5em}
\caption[]{The proposed CNN architecture. After a series of 7 convolutional layers, a fully-connected layer segues to two heads: one for regressing the sun position, and another one for the sky and camera parameters. The ELU activation function~\cite{clevert-iclr-16} is used on all layers except the outputs. }
\label{fig:cnn-architecture}
\end{figure}

\subsection{Validation of the optimization procedure}

While our fitting procedure minimizes reconstruction errors w.r.t.\ the panorama pixel intensities, the radiometrically uncalibrated nature of this data means that these fits may not accurately represent the true lighting conditions. We validate the procedure in two ways. First, the sun position estimation algorithm is evaluated on 543 panoramic sky images from the Laval HDR sky database~\cite{hdrdb,lalonde-3dv-14}, which contains ground truth sun position, and which we tonemapped and converted to JPG to simulate the conditions in SUN360. The median sun position estimation error of this algorithm is 4.59\degree ~(25th prct. = 1.96\degree, 75th prct. = 8.42\degree). Second, we ask a user to label 1,236 images from the SUN360 dataset, by indicating whether the estimated sky parameters agree with the scene visible in the panorama. To do so, we render a bunny model on a ground plane, and light it with the sky synthesized by the physical model. We then ask the user to indicate whether the bunny is lit similarly to the other elements present in the scene. In all, 65.6\% of the images were deemed to be a successful fit, which is testament to the challenging imaging conditions present in the dataset.

\section{Learning to predict outdoor lighting}
\label{sec:cnn}

\subsection{Dataset organization}

To train the CNN, we first apply the optimization procedure from sec.~\ref{sec:optimization} to 38,814 high resolution outdoor panoramas in the SUN360~\cite{xiao-cvpr-12} database. We then extract 7 photos from each panorama using a standard pinhole camera model and randomly sampling its parameters: its elevation with respect to the horizon in $[-20^\circ, 20^\circ]$, azimuth in $[-180^\circ, 180^\circ]$, and vertical field of view in $[35^\circ, 68^\circ]$. The resulting photos are bilinearly interpolated from the panorama to a resolution $320 \times 240$, and used directly to train the CNN described in the next section. This results in a dataset of 271,698 pairs of photos and their corresponding lighting parameters, which is split into (261,288 / 1,751 / 8,659) subsets for (train / validation / test). These splits were computed on the panoramas to ensure that photos taken from the same panorama do not end up in training and test. Example panoramas and corresponding photos are shown in fig.~\ref{fig:evaluation_example_sun_position}. 

\subsection{CNN architecture}

We adopt a standard feed-forward convolutional neural network to learn the relationship between the input image $I$ and the lighting parameters. As shown in fig.~\ref{fig:cnn-architecture}, its architecture is composed of 7 convolutional layers, followed by a fully-connected layer. It then splits into two separate heads: one for estimating the sun position (left in fig.~\ref{fig:cnn-architecture}), and one for the sky and camera parameters (right in fig.~\ref{fig:cnn-architecture}). 

The sun position head outputs a probability distribution over the likely sun positions $\mathbf{s}$ by discretizing the sky hemisphere into 160 bins (5 for elevation, 32 for azimuth), and outputs a value for each of these bins. This was also done in~\cite{lalonde-ijcv-12}. As opposed to regressing the sun position directly, this has the advantage of indicating other regions believed to be likely sun positions in the prediction, as illustrated in fig.~\ref{fig:evaluation_example_sun_position} below. The parameters head directly regresses a 4-vector of parameters $\mathbf{q}$: 2 for the sky ($\omega$, $t$), and 2 for the camera (elevation and field of view). The ELU activation function~\cite{clevert-iclr-16} and batch normalization~\cite{ioffe-jmlr-15} are used at the output of every layer. 

\subsection{Training details}

We define the loss to be optimized as the sum of two losses, one for each head: 
\begin{equation}
\mathcal{L}(\mathbf{s}^*, \mathbf{q}^*, \mathbf{s}, \mathbf{q}) = \mathcal{L}(\mathbf{s}^*, \mathbf{s}) + \beta \mathcal{L}(\mathbf{q}^*, \mathbf{q}) \,,
\label{eqn:loss}
\end{equation}
where $\beta = 160$ to compensate for the number of bins in $\mathbf{s}$. The target sun position $\mathbf{s}^*$ is computed for each bin $\mathbf{s}_j$ as 
\begin{equation}
\mathbf{s}^*_j = \exp(\kappa \mathbf{l}_s^{*\mathsf{T}} \mathbf{l}_j) \,,
\label{eqn:vmf}
\end{equation}
and normalized so that $\sum_j \mathbf{s}^*_j = 1$. The equation in (\ref{eqn:vmf}) represents a von Mises-Fisher distribution~\cite{banerjee-jmlr-05} centered about the ground truth sun position $\mathbf{l}_s$. Since the network must predict a confident value around the sun position, we set $\kappa = 80$. The target parameters $\mathbf{q}^*$ are simply the ground truth sky and camera parameters. 

\begin{figure}[t]
    \centering
    \footnotesize
    \setlength{\tabcolsep}{1pt}
    \begin{tabular}{cc}
    \multicolumn{2}{c}{\includegraphics[width=0.6\linewidth]{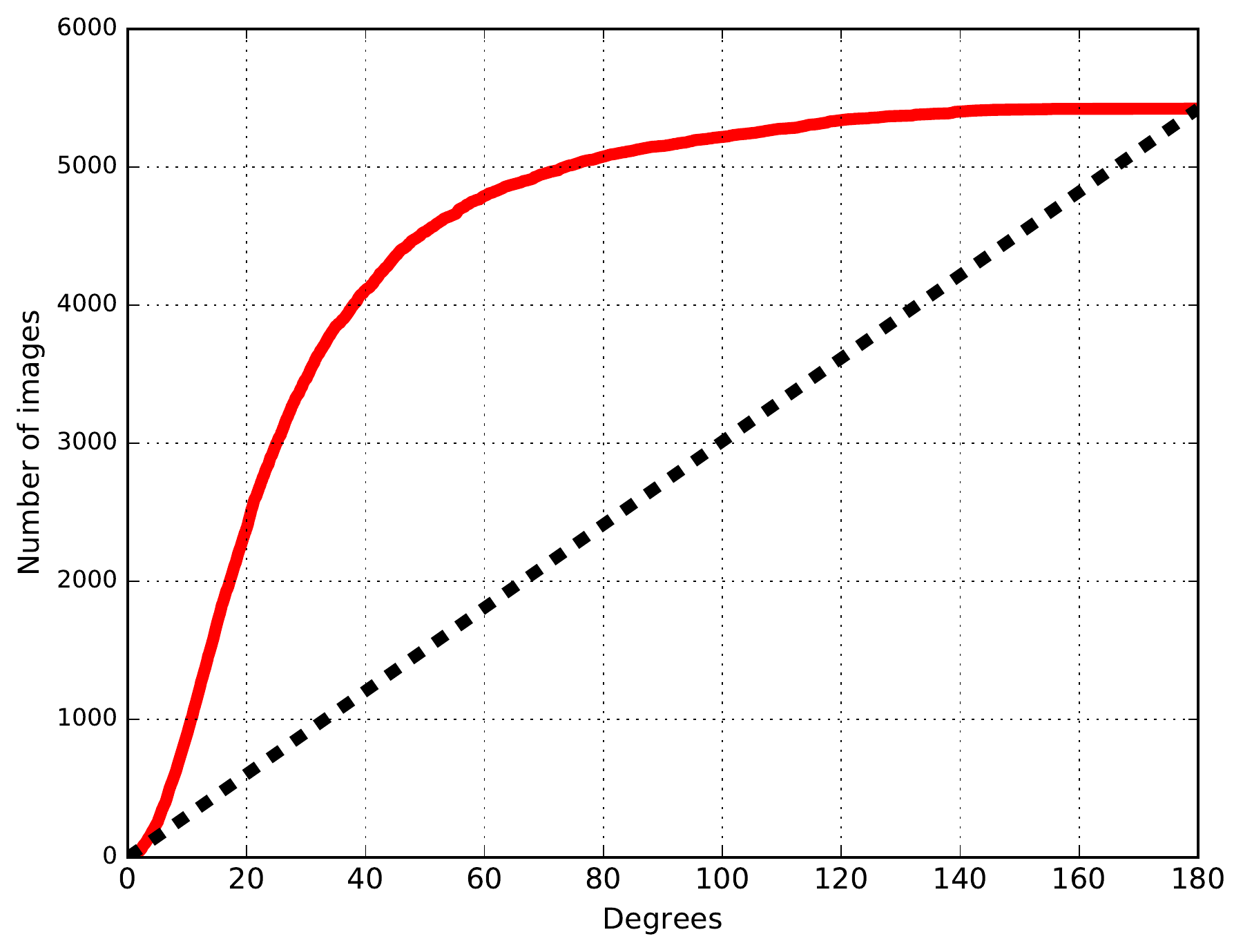}} \\
    \multicolumn{2}{c}{(a) Angular error} \\
    \includegraphics[width=0.48\linewidth]{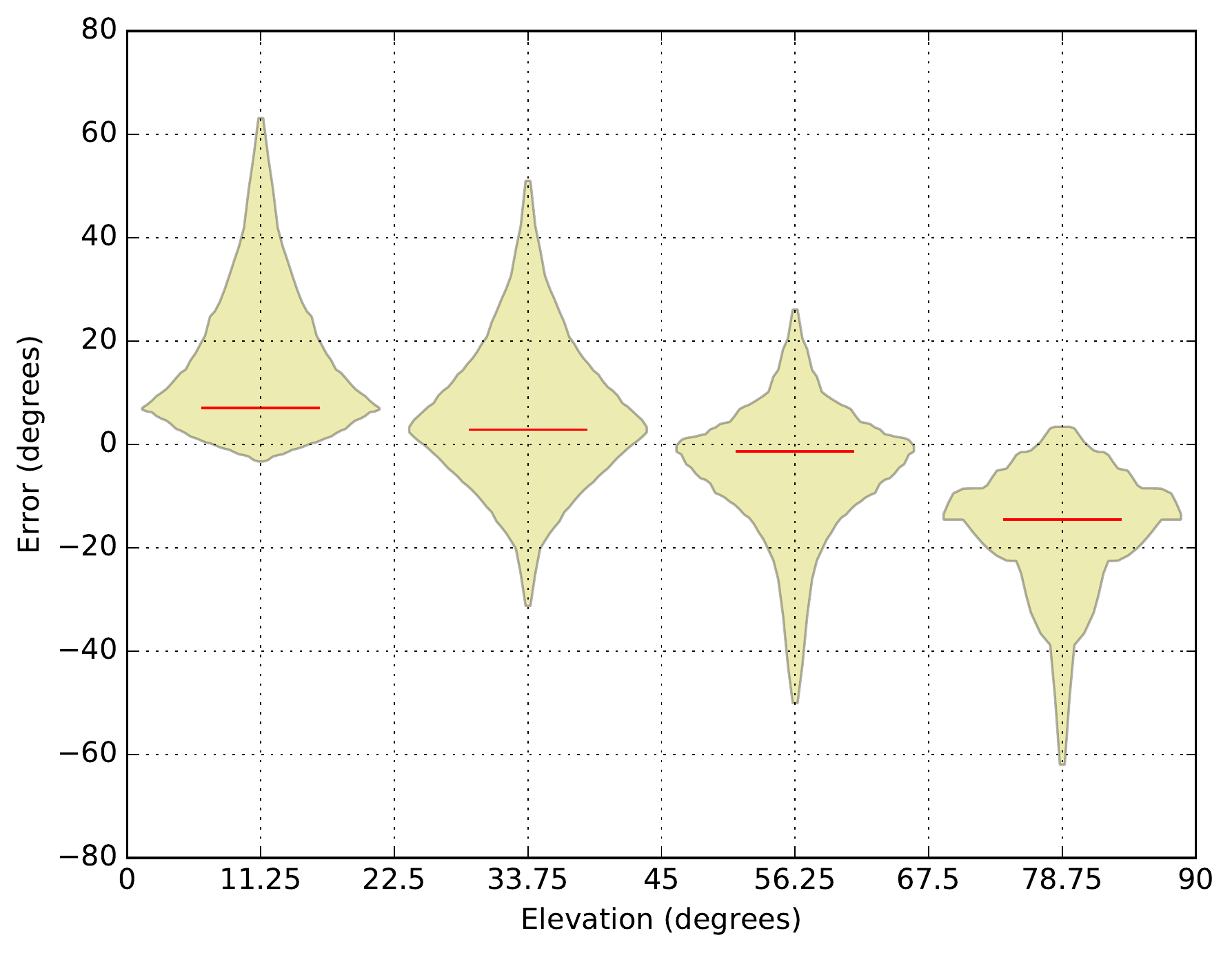} & 
    \includegraphics[width=0.48\linewidth]{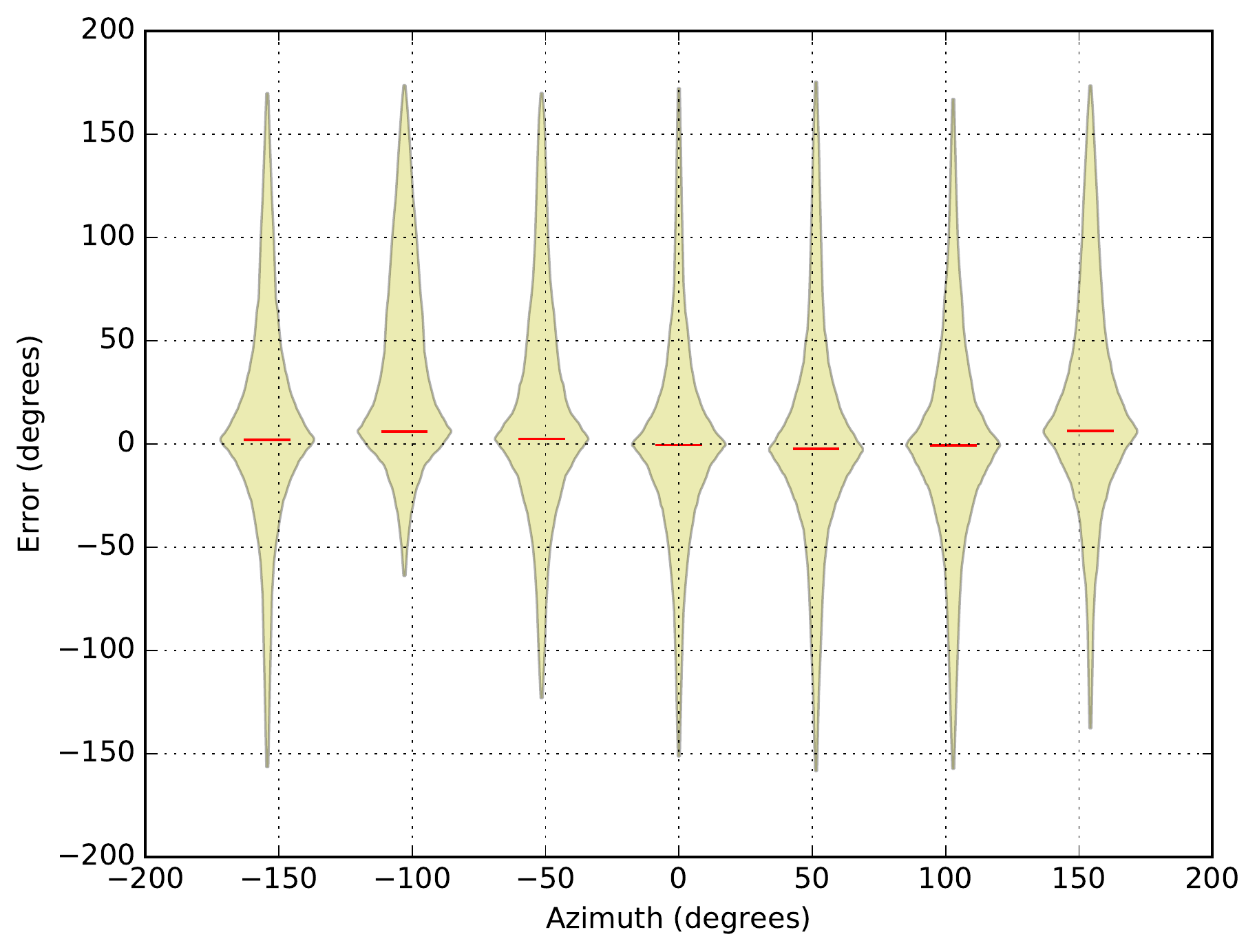} \\
    (b) Elevation error & (c) Azimuthal error
    \end{tabular}
    \vspace{.5em}
    \caption{Quantitative evaluation of sun position estimation on all 8659 images in the SUN360 test set. (a) The cumulative distribution function of the angular error on the sun position. The estimation error as function of the sun elevation (b) and (c) azimuth relative to the camera (0\degree ~means the sun is in front of the camera). The last two figures are displayed as ``box-percentile plots''~\cite{esty-jss-03}, where the envelope of each bin represents the percentile and the median is shown as a red bar.
    }
    \label{fig:evaluation_performance_sun_position}
\end{figure} 

\begin{figure}[!th]
    \centering
    \footnotesize
    \setlength{\tabcolsep}{1pt}
    \begin{tabular}{cc}
    \includegraphics[height=3.2cm]{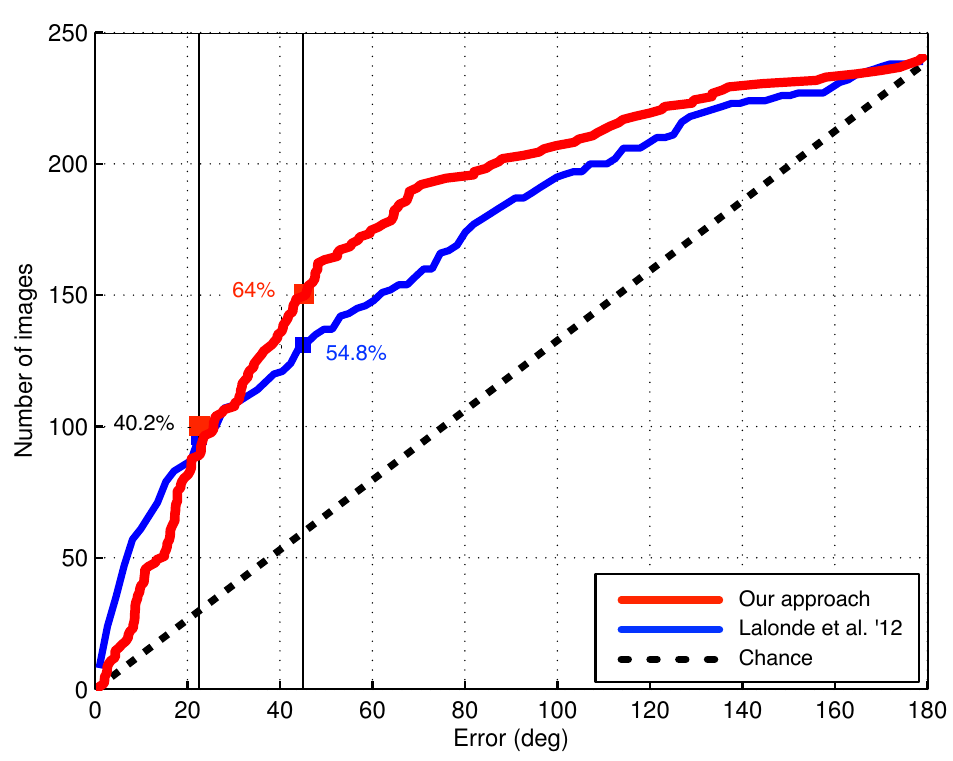} & 
    \includegraphics[height=3.2cm]{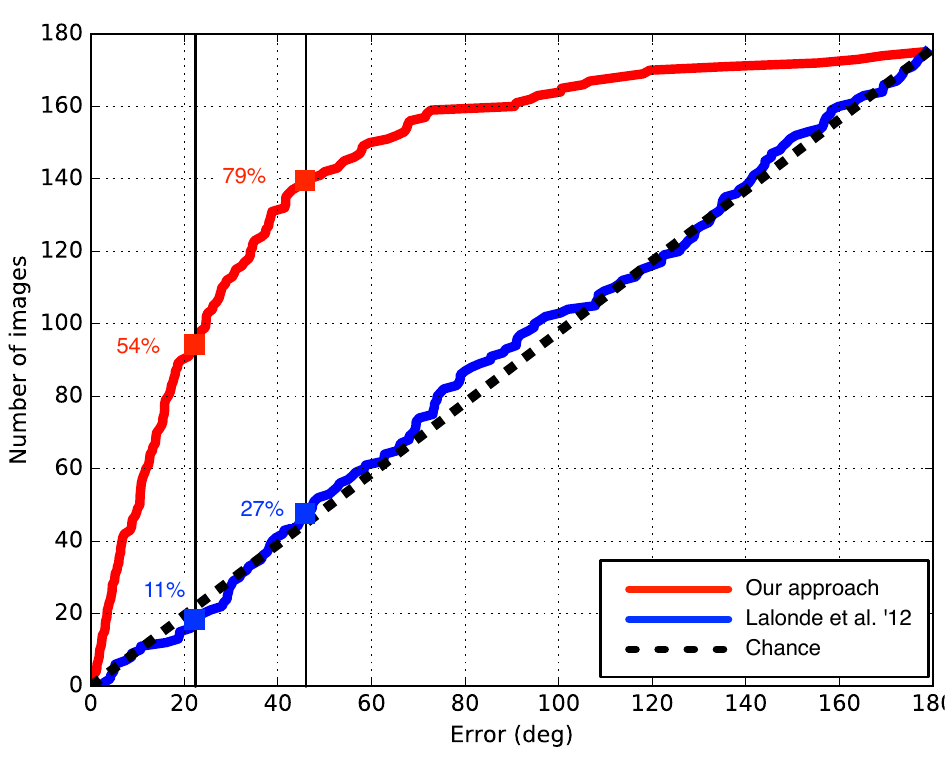} \\
    (a) Dataset from~\cite{lalonde-ijcv-12} &
    (b) Subset of SUN360 test set
    \end{tabular}
    \vspace{.5em} 
    \caption{Comparison with the method of Lalonde et al.~\cite{lalonde-ijcv-12} showing the cumulative sun azimuth estimation error on (a) their original dataset, and (b) a 176-image subset from the SUN360 test set. (a) While our method has similar error in an octant (less than $22.5^\circ)$, the precision in a quadrant (less than $45^\circ$) significantly improves by approximately 10\%. (b) The 176-images SUN360 test subset contains much more challenging images where methods based on the detection of explicit cues (as in~\cite{lalonde-ijcv-12}) fail. Our deep learning based approach remains robust and achieves high performance on both datasets.}
    \label{fig:comparison-lalonde12}
\end{figure}

We use a MSE loss for $\mathcal{L}(\mathbf{q}^*, \mathbf{q})$, and a Kullback-Leibler (KL) divergence loss for the sun position $\mathcal{L}(\mathbf{s}^*, \mathbf{s})$. Using the KL divergence is needed because we wish the network to learn a \emph{distribution} over the sun positions, rather than the most likely position. 

The loss in (\ref{eqn:loss}) is minimized via stochastic gradient descent using the Adam optimizer~\cite{kingma-iclr-15} with an initial learning rate of $\eta=0.01$. Training is done on mini-batches of 128 exemplars, and regularized via early stopping. The process typically converges in around 7--8 epochs, because our CNN is not as deep as most modern feed-forward CNN used in vision. Moreover, the high initial learning rate used combined with our large dataset further helps in reducing the number of epochs required for training.

\newcommand{\EvalHeight}{3.5cm}
\newcommand{\RndrHght}{3.2cm}
\newcommand{\RndrWdth}{4.1cm}
\newcommand{\CropWidth}{3.4cm}
\newcommand{\CropHght}{2.5cm}
\newcommand{\PanoWidth}{5.0cm}

\begin{figure*}
    \centering
    \includegraphics[width=\linewidth]{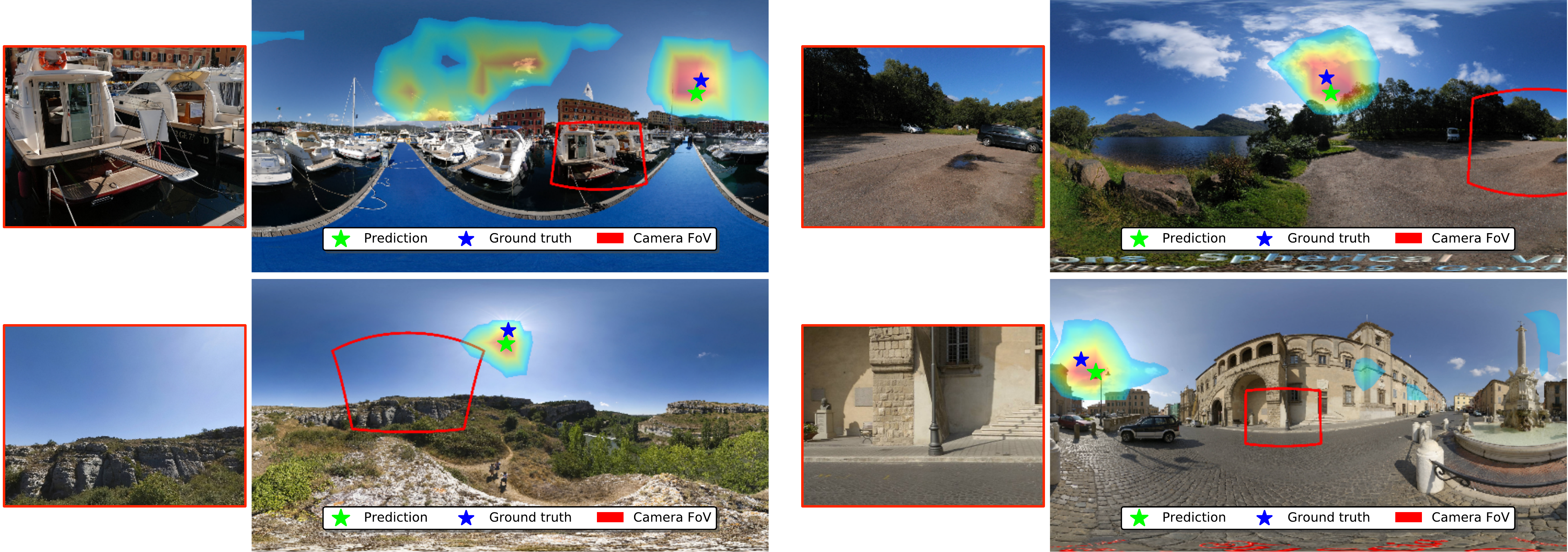}
    \caption{Examples of sun position estimation from a single outdoor image. For each example, the input image is shown on the left, and its corresponding location in the panorama is shown with a red outline. The color overlay displays the probability distribution of the sun position output by the neural network. A green star marks the most likely sun position estimated by the neural network, while a blue star marks the ground truth position. }
    \label{fig:evaluation_example_sun_position}
\end{figure*}

\section{Evaluation}
\label{sec:evaluation}

We evaluate the performance of the CNN at predicting the HDR sky environment map from a single image in a variety of ways. First, we present how well the network does at estimating the illumination parameters on the SUN360 dataset. We then show how virtual objects relit by the \emph{estimated} environment maps differ from their renders obtained with the ground truth parametric model, still on the SUN360. Finally, we acquired a small set of HDR outdoor panoramas, and compare our relighting results with those obtained with actual HDR environment maps. 

\subsection{Illumination parameters on SUN360}

\paragraph{Sun position}

We begin by evaluating the performance of the CNN at predicting the sun position from a single input image. Fig.~\ref{fig:evaluation_performance_sun_position} shows the quantitative performance at this task using three plots: the cumulative distribution function of sun angular estimation error, and detailed error histograms for each of the elevation and azimuth independently. We observe that 80\% of the test images have error less than 45\degree. Fig.~\ref{fig:evaluation_performance_sun_position}-(b) indicates that the network tends to underestimate the sun elevation in high elevation cases. This may be attributable to a lack of such occurrences in the training dataset---high sun elevations only occur between the tropics, and at specific times of year because of the Earth's tilted rotation axis. Fig.~\ref{fig:evaluation_performance_sun_position}-(c) shows that the CNN is not biased towards an azimuth position, and is robust across the entire range. Fig.~\ref{fig:evaluation_example_sun_position} shows examples of our sun position predictions overlayed over the panoramas that the test images were cropped from. Note that our method is able to accurately predict the sun direction across a wide range of scenes, field of views, and layouts.

We quantitatively compare our approach to that of \cite{lalonde-ijcv-12} at the task of sun azimuth estimation from a single image. Results are reported in fig.~\ref{fig:comparison-lalonde12}. First, fig.~\ref{fig:comparison-lalonde12}-(a) shows a comparison of both approaches on the 239-image dataset of \cite{lalonde-ijcv-12}. While our method has similar error in an octant (less than 22.5\degree), the precision in a quadrant (less than 45\degree) is significantly improved (by approximately 10\%) by our CNN-based approach. Fig.~\ref{fig:comparison-lalonde12}-(b) shows the same comparison on a 176-image subset of the SUN360 test set used in this paper. In this case, the approach of Lalonde et al.~\cite{lalonde-ijcv-12} fails while the CNN reports robust performance, comparable to fig.~\ref{fig:comparison-lalonde12}-(a). This is probably due to the fact that the SUN360 test set contains much more challenging images that are often devoid of strong, explicit illumination cues. These cues, which are expressly relied upon by \cite{lalonde-ijcv-12}, are critical to the success of such methods.
\vspace{-1.5em}
\paragraph{Turbidity and exposure}

\begin{figure}[!th]
    \centering
    \footnotesize
    \setlength{\tabcolsep}{1pt}
    \begin{tabular}{cc}
    \includegraphics[height=3.1cm]{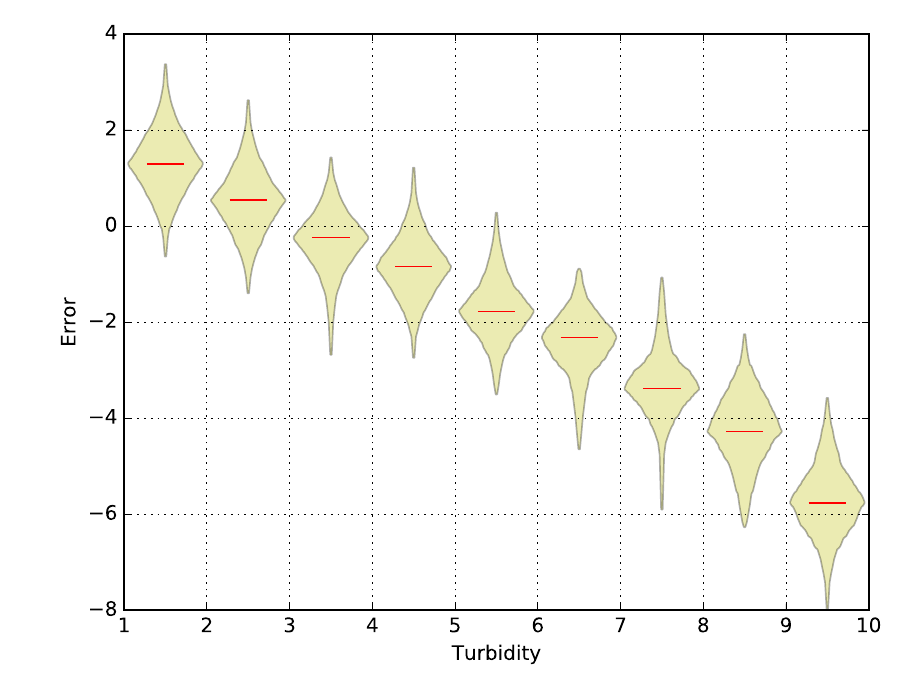} &
    \includegraphics[height=3.1cm]{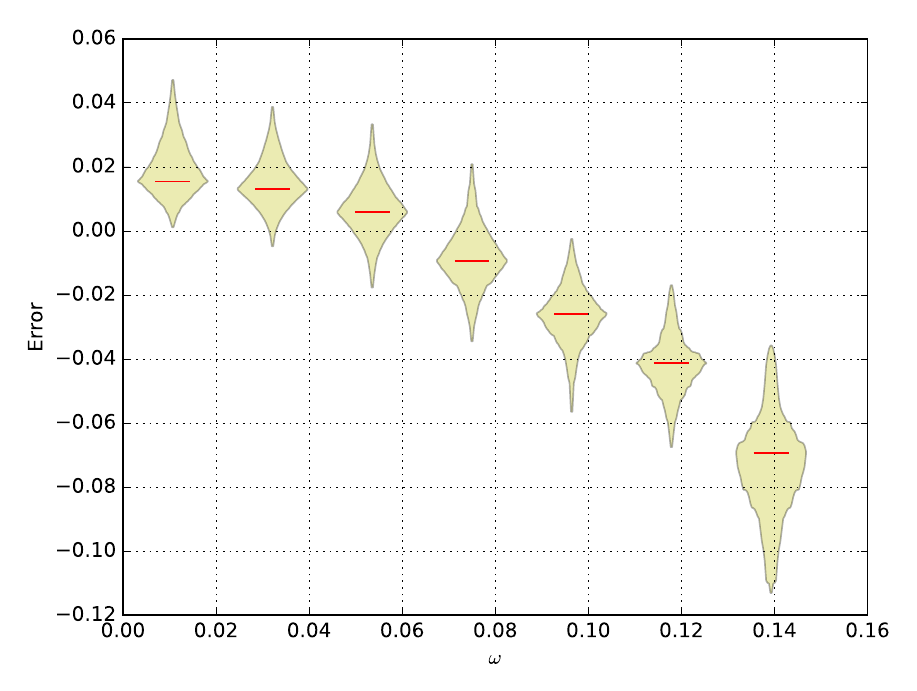} \\
    (a) Turbidity $t$ &
    (b) Exposure $\omega$ 
    \end{tabular}
    \vspace{.25em}
    \caption{Quantitative evaluation for turbidity $t$ and exposure $\omega$. The distribution of errors are displayed as ``box-percentile'' plots (see fig.~\ref{fig:evaluation_performance_sun_position}). The CNN tends to favor clear skies (low turbidity), and has higher errors when the exposure is high.}
    \label{fig:evaluation_sky_parameters}
\end{figure}

\begin{figure}[!t]
\centering
\footnotesize
\includegraphics[width=\linewidth]{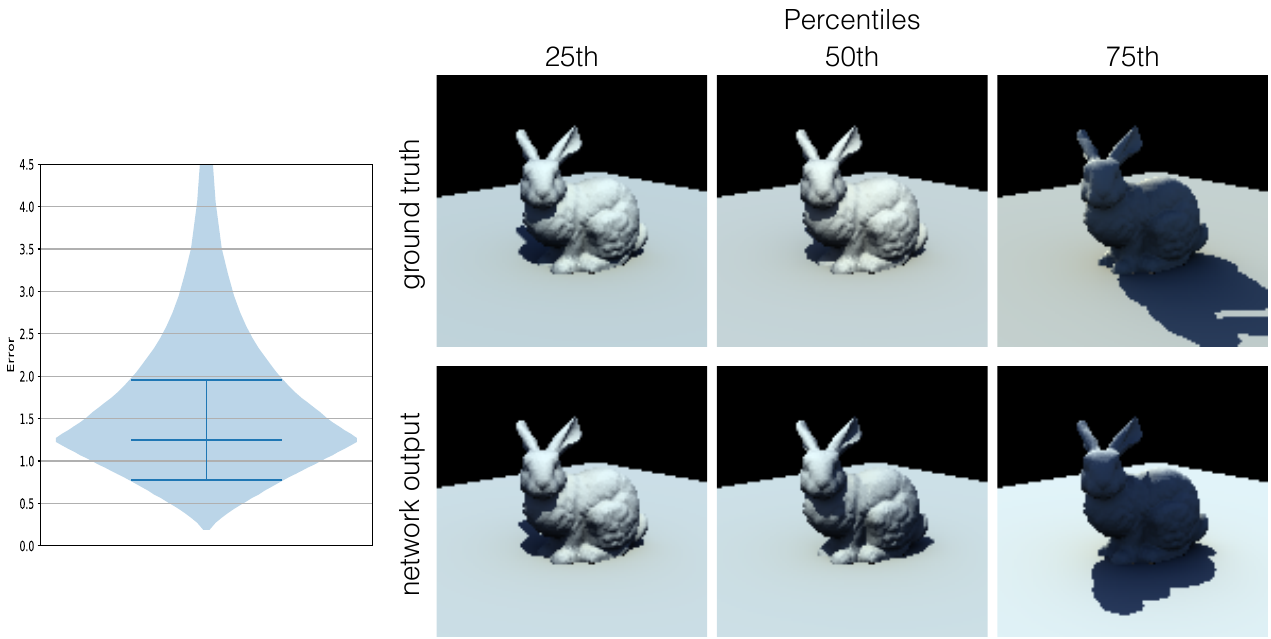} \\
(a) RMSE
\includegraphics[width=\linewidth]{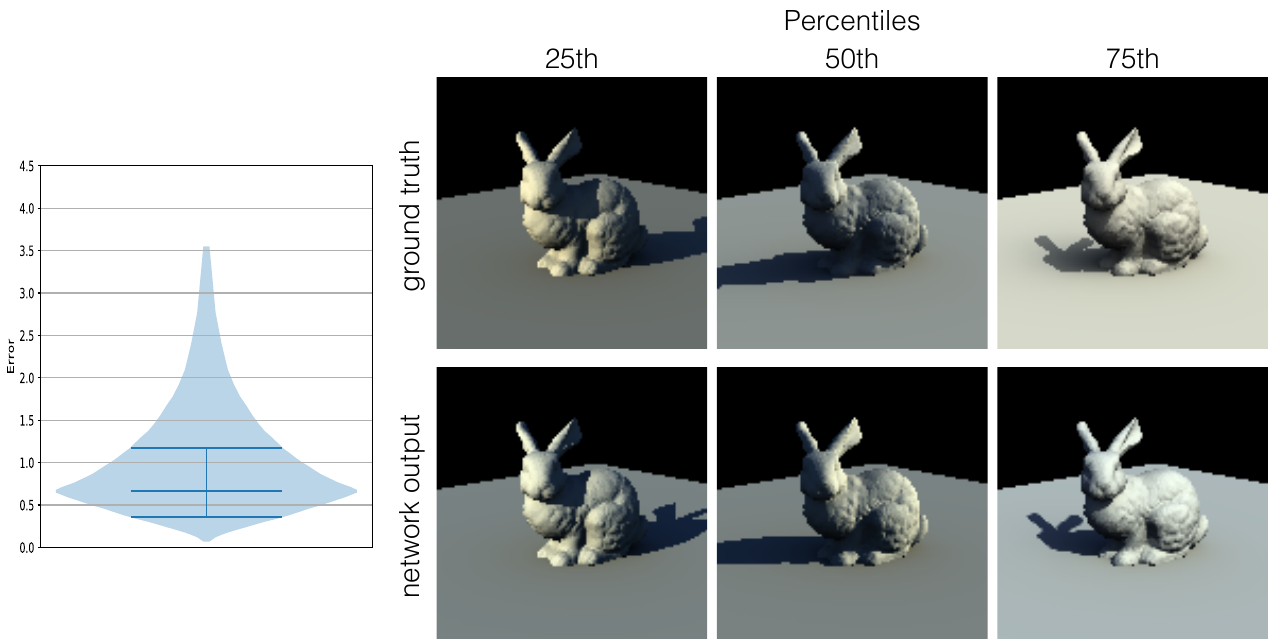} \\
(b) Scale-invariant RMSE
\includegraphics[width=\linewidth]{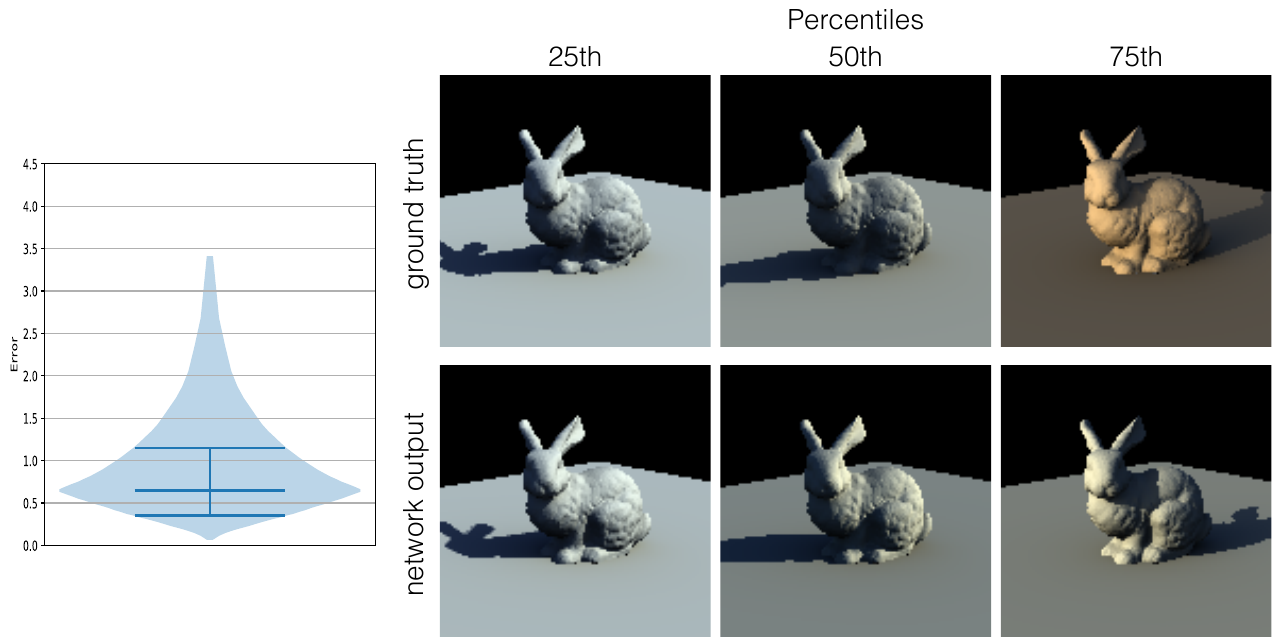} \\
(c) Per-color scale-invariant RMSE
\vspace{.25em}
\caption{Quantitative relighting comparison with the ground truth lighting parameters on the SUN360 dataset. We compute three types of error metrics: (a) RMSE, (b) scale-invariant RMSE~\cite{grosse-iccv-09}, and (c) per-color scale-invariant RMSE. The plots on the left shows the distribution of errors with the median, 25th and 75th percentiles identified with blue bars. For each measure, examples corresponding to particular error levels are shown to give a qualitative sense of performance. Renders obtained with the ground truth (estimated) lighting parameters are shown in the top (bottom) row.}
\label{fig:quantitative-relighting-sun360}
\vspace{-1em}
\end{figure}


We evaluate the regression performance for the turbidity $t$ and exposure $\omega$ lighting parameters on the SUN360 test set, and report the results in fig.~\ref{fig:evaluation_sky_parameters}. Overall, the network tends to favor low turbidity estimates of the sky (as the dataset contains a majority of such examples). In addition, the network successfully estimates low exposure values, but has a tendency to underestimate images with high exposures.
\vspace{-1.5em}
\paragraph{Camera parameters}

A detailed performance analysis is available in the supplementary material. In a nutshell, the CNN achieves error of less than 7\degree ~for the elevation and 11\degree ~in field of view for 80\% of the test images. 




\begin{figure}[!th]
    \centering
    \footnotesize
    \setlength{\tabcolsep}{1pt}
    \begin{tabular}{cc}
    \includegraphics[width=.47\linewidth]{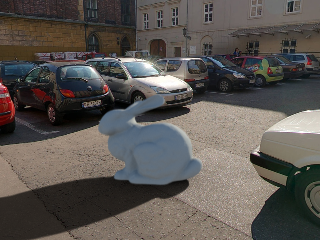} &
    \includegraphics[width=.47\linewidth]{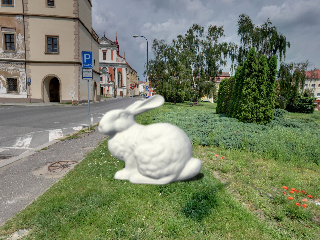} \\
    \includegraphics[width=.47\linewidth]{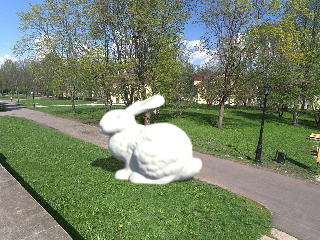} &
    \includegraphics[width=.47\linewidth]{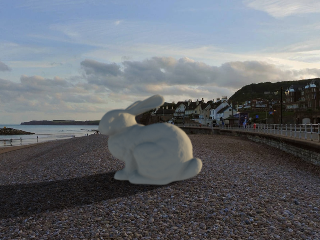}
    \end{tabular}
    \caption{Virtual object insertion with automated lighting estimation. From a single image, the CNN predicted a full HDR sky map, which is used to render an object into the image. No additional steps are required. More results on automated object insertion are available in the supplementary materials.}
    \label{fig:evaluation_render_examples}
\end{figure}

\begin{figure}[!th]
    \centering
    \footnotesize
    \setlength{\tabcolsep}{1pt}
    \begin{tabular}{ccc}
    \includegraphics[width=.47\linewidth]{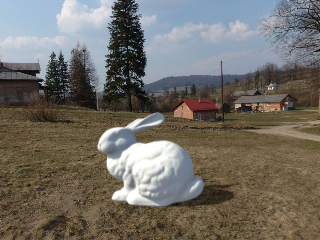} & 
    \includegraphics[width=.47\linewidth]{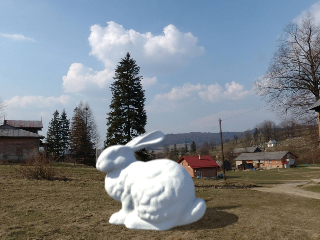} \\
    Estimated elevation: -9\degree & 
    Estimated elevation: 3.5\degree & 
    \end{tabular}
    \vspace{.25em}
    \caption{Virtual object insertion with automated lighting and camera elevation estimation. The two images are taken at the same location with the camera pointing downwards (left) and upwards (right). The elevation of the virtual camera used to render the bunny model is set to the value predicted by the CNN, resulting in a bunny which realistically rests on the ground. }
    \label{fig:evaluation-elevation}
    \vspace{-1em}
\end{figure}

\subsection{Relighting on SUN360}

Another way of evaluating the performance is by comparing the appearance of a Lambertian 3D model rendered with the estimated lighting, with that of the same model lit by the ground truth. Fig.~\ref{fig:quantitative-relighting-sun360} provides such a comparison, by showing three different error metrics computed on renderings obtained on our test set. The error metrics are the (a) RMSE, (b) scale-invariant RMSE, and (c) per-color scale-invariant RMSE. The scale-invariant versions of RMSE are defined similarly to Grosse et al.~\cite{grosse-iccv-09}, except that the scale factor is computed on the entire image (instead of locally as in~\cite{grosse-iccv-09}). The ``per-color'' variant computes a different scale factor for each color channel to mitigate differences in white balance. The black background in the renders is masked out before computing the metrics.
  
To give a sense of what those numbers mean qualitatively, fig.~\ref{fig:quantitative-relighting-sun360} also provides examples corresponding to each of the (25, 50, 75)th error percentiles. Even examples in the 75th error percentile look good qualitatively. Slight differences in the sun direction and the overall color can be observed, but they still lie within reasonable limits.

Fig.~\ref{fig:evaluation_render_examples} shows examples of virtual objects inserted into images after being rendered with our estimated HDR illumination. As these examples show, our technique is able to infer plausible illumination conditions ranging from sunny to overcast, and high noon to dawn/dusk, resulting in natural-looking composite images. Fig.~\ref{fig:evaluation-elevation} shows that the camera elevation estimated from the CNN can be used within the rendering pipeline to automatically rotate the virtual camera used to render the object. In these results, a simple ground plane is used to model the interactions between the virtual object and its environment, and the object is placed manually at a fixed distance in front of the camera.

\begin{figure*}[t]
    \centering 
    \footnotesize
    \setlength{\tabcolsep}{1pt}
    \begin{tabular}{cccccc}
    Ground truth & Estimated & Ground truth & Estimated & Ground truth & Estimated \\
    \includegraphics[width=.16\linewidth]{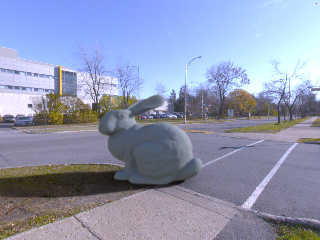} &
    \includegraphics[width=.16\linewidth]{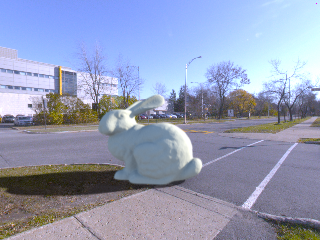}\: &
    \includegraphics[width=.16\linewidth]{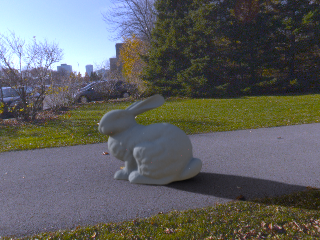} &
    \includegraphics[width=.16\linewidth]{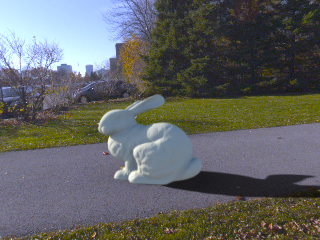}\: &
    \includegraphics[width=.16\linewidth]{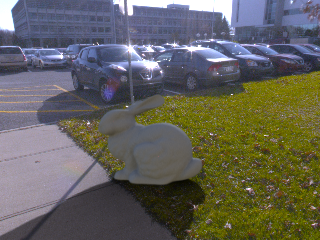} &
    \includegraphics[width=.16\linewidth]{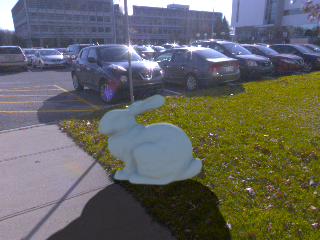} \\
    \multicolumn{2}{c}{
    \includegraphics[width=\RndrWdth]{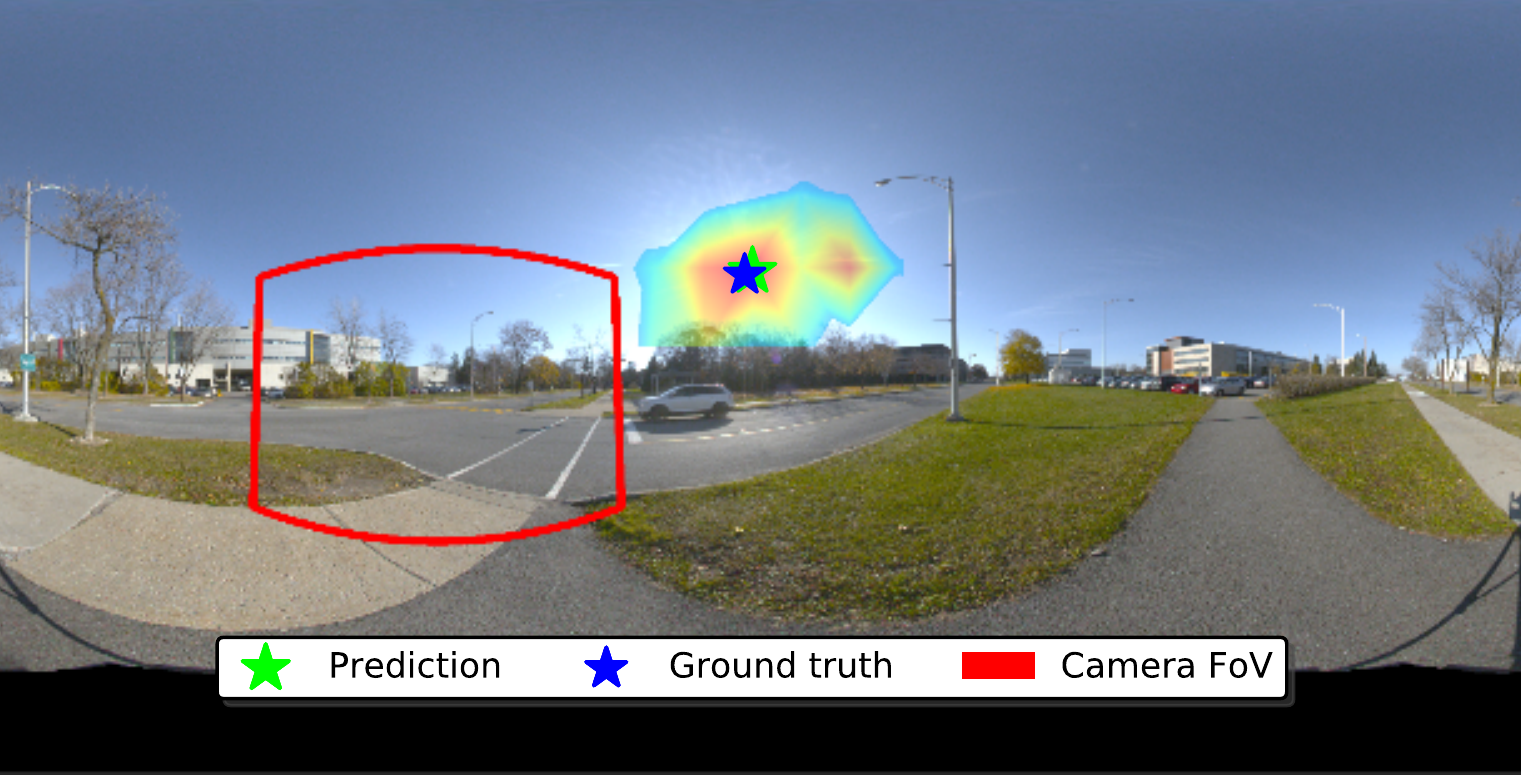}} &
    \multicolumn{2}{c}{
    \includegraphics[width=\RndrWdth]{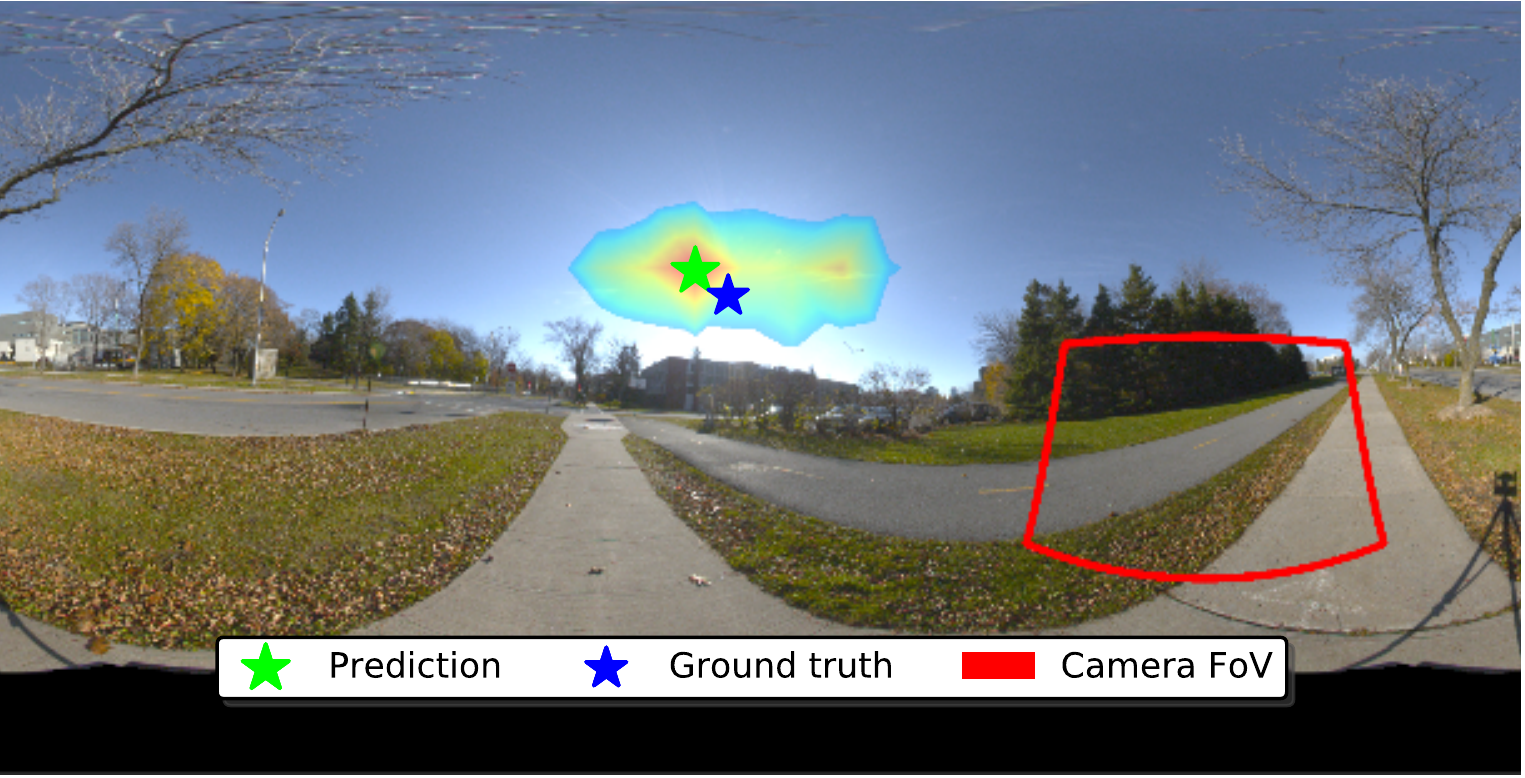}} &
    \multicolumn{2}{c}{
    \includegraphics[width=\RndrWdth]{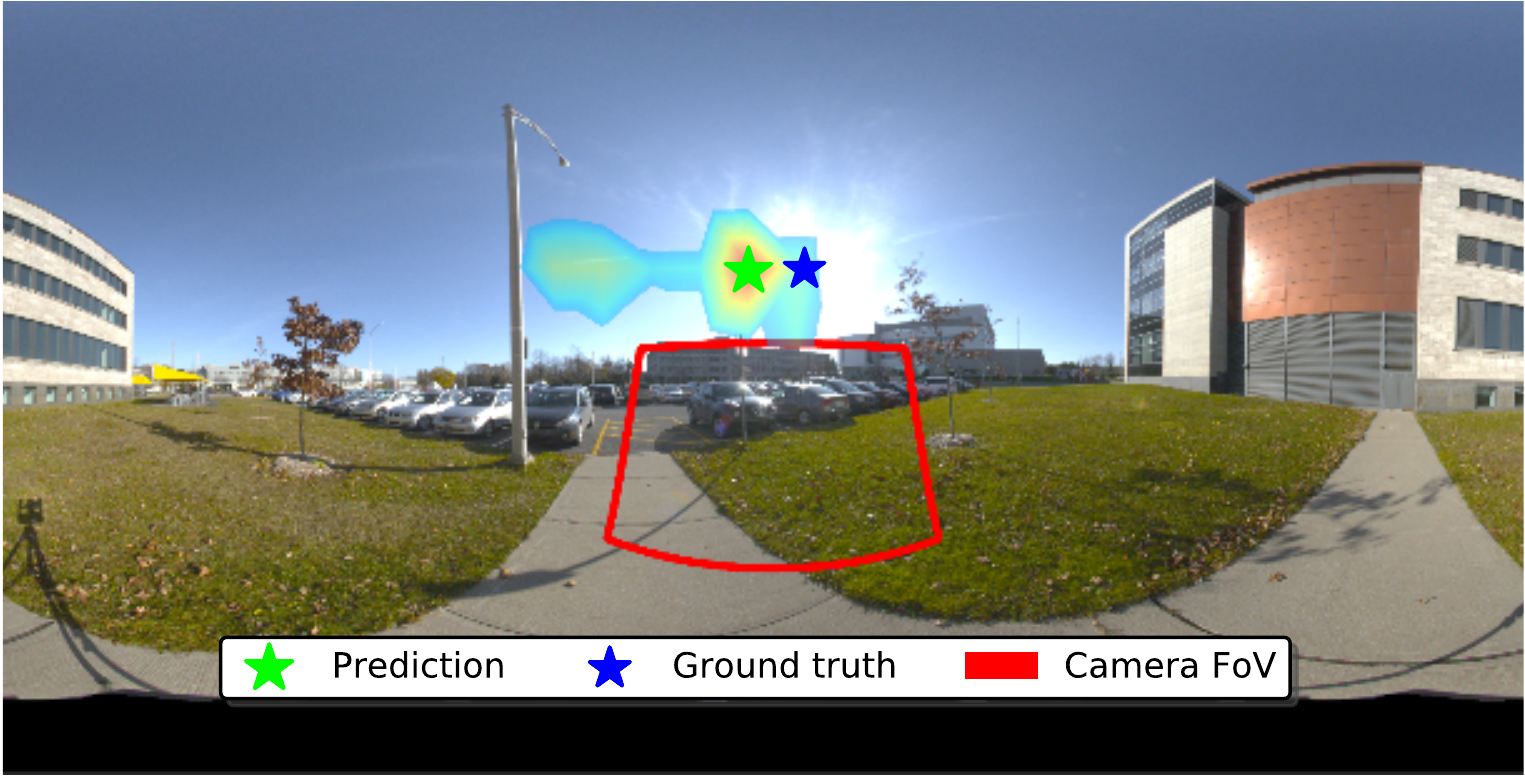}} \\
    \multicolumn{2}{c}{(a)} & \multicolumn{2}{c}{(b)} & \multicolumn{2}{c}{(c)}
    \end{tabular}
    \caption{Object relighting comparison with ground truth illumination conditions on captured HDR panoramas. For each example, the top row shows (left) a bunny model relit by the ground truth HDR illumination conditions captured in situ; (right) the same bunny model, relit by the illumination conditions estimated by the CNN solely from the background image, completely automatically. No further adjustment (e.g. overall brightness, saturation, etc.) was performed. The bottom row shows the original environment map, field of view of the camera (in red), and the distribution on sun position estimation (as in fig.~\ref{fig:evaluation_example_sun_position}). Please see additional results on our project page.}
    \label{fig:hdr-panoramas-validation}
    \vspace{-1em}
\end{figure*}

\subsection{Validation with HDR panoramas}
\label{subsec:relighting_hdr}

To further validate our approach, we captured a small dataset of 19 unsaturated, outdoor HDR panoramas. To properly expose the extreme dynamic range of outdoor lighting, we follow the approach proposed by Stumpfel et al.~\cite{stumpfel-afrigraph-04}. We captured 7 bracketed exposures ranging from 1/8000 to 8 seconds at f/16, using a Canon EOS 5D Mark III camera installed on a tripod, and fitted with a Sigma EXDG 8mm fisheye lens. A 3.0 ND filter was installed behind the lens, necessary to accurately measure the sun intensity. The exposures were stored as 14-bit RAW images at full resolution. The process was repeated at 6 azimuth angles by increments of 60\degree ~to cover the entire 360\degree ~panorama. The resulting 42 images were fused using the \textsc{PTGui} commercial stitching software. To facilitate the capture process, the camera was mounted on a programmable robotic tripod head, allowing for repeatable and precise capture. 


To validate the approach, we extract limited field of view photos from the HDR panoramas and save them as JPEG files. The CNN is then applied to the input photos to predict their illumination conditions. Then, we compare relighting results obtained by rendering a bunny model with: 1) the HDR panorama itself, which represents the ground truth lighting conditions; and 2) the estimated lighting conditions. Example results are shown in fig.~\ref{fig:hdr-panoramas-validation}. While we note that the exposure $\omega$ is slightly overestimated (resulting in a render that is brighter than the ground truth), the relit bunny appears quite realistic. 
  
\begin{figure}
    \centering
    \includegraphics[width=\linewidth]{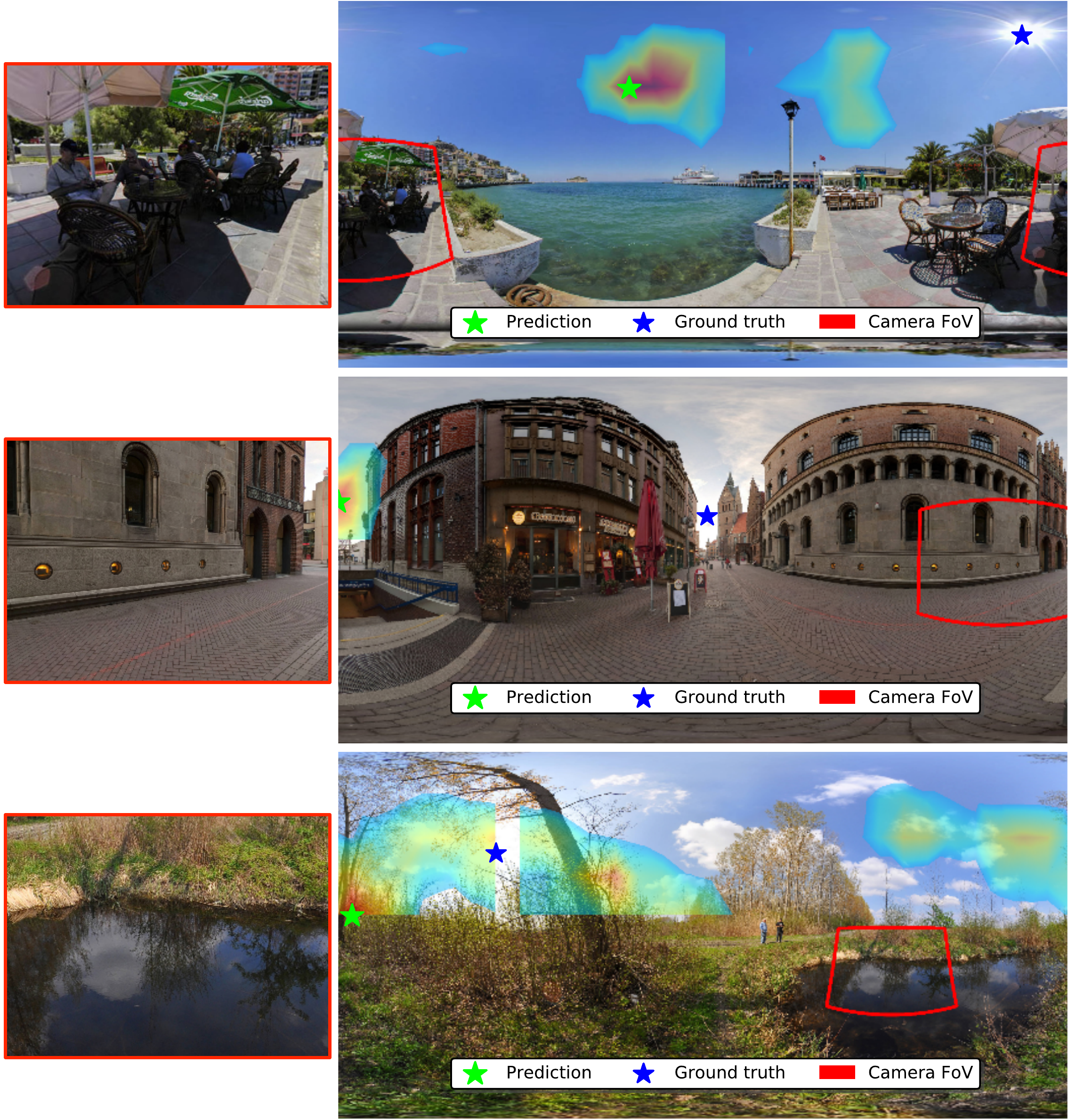}
    \caption{Typical failure cases of sun position estimation from a single outdoor image. See fig.~\ref{fig:evaluation_example_sun_position} for an explanation of the annotations. Failure cases occur when illumination cues are mixed with complex geometry (top), absent from the image (middle),  or in the presence of mirror-like surfaces (bottom).}
    \label{fig:evaluation_example_sun_position_failure_cases}
    \vspace{-2em}
\end{figure}

\section{Discussion}

In this paper, we propose what we believe to be the first end-to-end approach to automatically predict full HDR lighting models from a single outdoor LDR image of a general scene, which can readily be used for image-based lighting. Our key idea is to train a deep CNN on pairs of photos and panoramas in the SUN360 database, which we ``augment'' with HDR information via a physics-based model of the sky. We show that our method significantly outperforms previous work, and that it can be used to realistically insert virtual objects into photos.

Despite offering state-of-the-art performance, our method still suffers from some limitations. First, the Ho\v{s}ek-Wilkie sky model provides accurate representational accuracy for clear skies, but its accuracy degrades when cloud cover increases as the turbidity $t$ is not enough to model completely overcast situations as accurately as for clear skies. Optimizing its parameters on overcast panoramas often underestimates the turbidity, resulting in a bias toward low turbidity in the CNN. We are currently investigating ways of mitigating this issue by combining the HW model with another sky model, better-suited for overcast skies. Another limitation is that the resulting environment map models the sky hemisphere only. While this does not affect diffuse objects such as the bunny model used in this paper, it would be more problematic for rendering specular materials, as none of the scene texture would be reflected off its surface. It is likely that simple adjustments such as~\cite{khan-siggraph-06} could be helpful in making those renders more realistic.

\section{Acknowledgments}

The authors would like to thank Marc-André Gardner for his help with the architecture and optimization. Parts of this work were done while Yannick Hold-Geoffroy was an intern at Adobe Research. This work was partially supported by the REPARTI Strategic Network, the FRQNT New Researcher Grant 2016NC189939 and the NSERC Discovery Grant RGPIN-2014-05314. We gratefully acknowledge the support of Nvidia with the donation of the GPUs used for this research.

{\small
\bibliographystyle{ieee}
\bibliography{main}
}

\end{document}